\documentclass{article}
\usepackage{CJKutf8}

\usepackage{graphicx}
\usepackage{arxiv}
\usepackage[T1]{fontenc}    
\usepackage{hyperref}       
\usepackage{url}            
\usepackage{booktabs}       
\usepackage{amsfonts}       
\usepackage{nicefrac}       
\usepackage{microtype}      
\usepackage{lipsum}
\usepackage[utf8]{inputenc}
\usepackage{multirow}
\usepackage{graphicx}

\title{TexSmart: A Text Understanding System for Fine-Grained NER and Enhanced Semantic Analysis}
\author{
  Haisong Zhang, Lemao Liu, Haiyun Jiang, Yangming Li, Enbo Zhao, \\
  \textbf{Kun Xu, Linfeng Song, Suncong Zheng, Botong Zhou, Jianchen Zhu, Xiao Feng,}\\
  \textbf{Tao Chen, Tao Yang, Dong Yu, Feng Zhang, Zhanhui Kang, Shuming Shi}\thanks{Project lead and chief architect}\\
  \\
  Tencent AI\\
 \texttt{\{texsmart, hansonzhang, redmondliu, shumingshi\}@tencent.com}
}

\begin{document}
\maketitle
\begin{CJK}{UTF8}{gbsn}

\begin{abstract}
This technique report introduces TexSmart, a text understanding system that supports fine-grained named entity recognition (NER) and enhanced semantic analysis functionalities.
Compared to most previous publicly available text understanding systems and tools, TexSmart holds some unique features.
First, the NER function of TexSmart supports over 1,000 entity types, while most other public tools typically support several to (at most) dozens of entity types.
Second, TexSmart introduces new semantic analysis functions like semantic expansion and deep semantic representation, that are absent in most previous systems.
Third, a spectrum of algorithms (from very fast algorithms to those that are relatively slow but more accurate) are implemented for one function in TexSmart, to fulfill the requirements of different academic and industrial applications.
The adoption of unsupervised or weakly-supervised algorithms is especially emphasized, with the goal of easily updating our models to include fresh data with less human annotation efforts.

The main contents of this report include major functions of TexSmart, algorithms for achieving these functions, how to use the TexSmart toolkit and Web APIs, and evaluation results of some key algorithms.

\end{abstract}


\section{Introduction}
The long-term goal of natural language processing (NLP) is to help computers understand natural language as well as we do, which is one of the most fundamental and representative challenges for artificial intelligence.
Natural language understanding includes a broad variety of tasks covering lexical analysis, syntactic analysis and semantic analysis.
In this report we introduce TexSmart\footnote{The English home-page of TexSmart is \url{https://texsmart.qq.com/en}}, a new text understanding system that provides enhanced named entity recognition (NER) and semantic analysis functionalities besides standard NLP modules.
Compared to most previous publicly-available text understanding systems \cite{loper2002nltk, OpenNLP, manning2014stanford, gardner-etal-2018-allennlp, che2010ltp, qiu2013fudannlp}, TexSmart holds the following key characteristics:

\begin{itemize}
  \item Fine-grained named entity recognition (NER)
  \item Enhanced semantic analysis
  \item A spectrum of algorithms implemented for one function, to fulfill the requirements of different academic and industrial applications
\end{itemize}
\begin{figure}[htb]
  \centering
  \includegraphics[width=\textwidth]
      {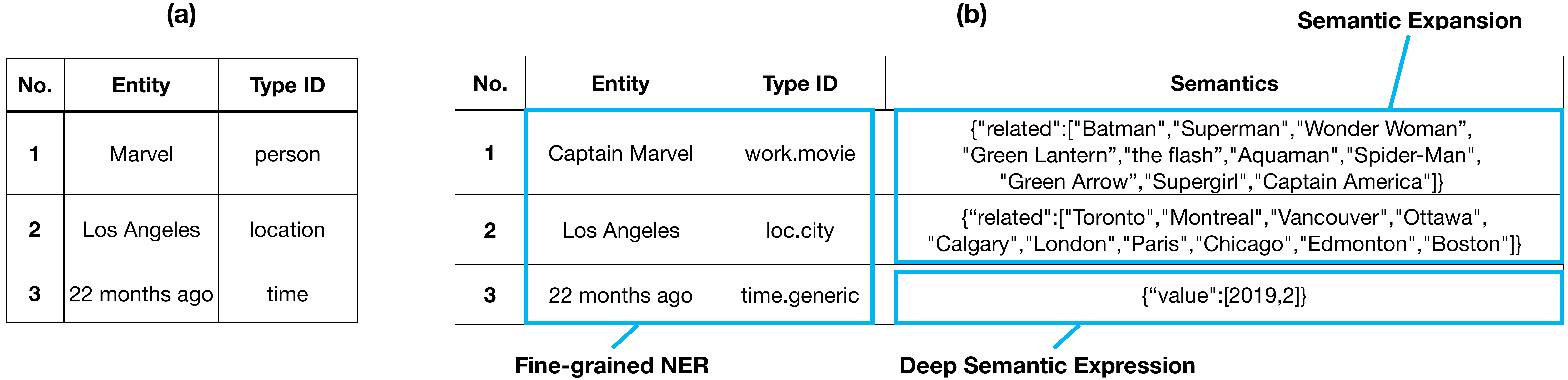}

 \caption{\label{fig:key-feat} Comparison between the NER results of a traditional text understanding system (in (a)) and the fine-grained NER and semantic analysis results provided by TexSmart (in (b)). The input sentence is ``Captain Marvel was premiered in Los Angeles 22 months ago.'' TexSmart successfully assigns fine-grained type ``work.movie'' to ``Captain Marvel'', while most traditional toolkits fail to do that. In addition, TexSmart gives more rich semantic analysis results for the input sentence.}
  \vspace{-0.1in}
\end{figure}

First, the fine-grained NER function of TexSmart supports over 1,000 entity types while most previous text understanding systems typically support several to (at most) dozens of coarse entity types (among which the most popular types are people, locations, and organizations).
Large-scale fine-grained entity types are expected to provide richer semantic information for downstream NLP applications.
Figure~\ref{fig:key-feat} shows a comparison between the NER results of a previous system and the fine-grained NER results of TexSmart.
It is shown that TexSmart recognizes more entity types (e.g., work.movie) and finer-grained ones (e.g., loc.city vs. the general location type).
Examples of entity types (and their important sub-types) which TexSmart is able to recognize include people, locations, organizations, products, brands, creative work, time, numerical values, living creatures, food, drugs, diseases, academic disciplines, languages, celestial bodies, organs, events, activities, colors, etc.

Second, TexSmart provides two advanced semantic analysis functionalities: semantic expansion, and deep semantic representation for a few entity types.
These two functions are not available in most previous public text understanding systems.
Semantic expansion suggests a list of related entities for an entity in the input sentence (as shown in Figure~\ref{fig:key-feat}). It provides more information about the semantic meaning of an entity. Semantic expansion could also benefit upper-layer applications like web search (e.g., for query suggestion) and recommendation systems.
For time and quantity entities, in addition to recognizing them from a sentence, TexSmart also tries to parse them into deep representations (as shown in Figure~\ref{fig:key-feat}).
This kind of deep representations is essential for some NLP applications.
For example, when a chatbot is processing query ``please book an air ticket to London at 4 pm the day after tomorrow'', it needs to know the exact time represented by ``4 pm the day after tomorrow''.

Third, a spectrum of algorithms is implemented for one task (e.g., part-of-speech tagging and NER) in TexSmart, to fulfill the requirements of different academic and industrial applications.
On one side of the spectrum are the algorithms that are very fast but not necessarily the best in accuracy. On the opposite side are those that are relatively slow yet delivering state-of-the-art performance in terms of accuracy.
Different application scenarios may have different requirements for efficiency and accuracy. Unfortunately, it is often very difficult or even impossible for a single algorithm to achieve the best in both speed and accuracy at the same time.
With multiple algorithms implemented for one task, we have more chances to better fulfill the requirements of more applications.

One design principle of TexSmart is to put a lot of efforts into designing and implementing unsupervised or weakly-supervised algorithms for a task, based on large-scale structured, semi-structured, or unstructured data.
The goal is to update our models easier to include fresh data with less human annotation efforts.

The rest part of this report is organized as follows: Major tasks and algorithms supported by TexSmart are described in Section 2. Then we demonstrate in Section 3 how to use the TexSmart toolkit and Web APIs. 
Finally, evaluation results of some algorithms on several key tasks are reported.

\section{System Modules}
Compared to most other public text understanding systems, TexSmart supports three unique modules, i.e., fine-grained NER, semantic expansion and deep semantic representation.
Besides, traditional tasks supported by both TexSmart and many other systems include word segmentation, part-of-speech (POS) tagging, coarse-grained NER, constituency parsing, semantic role labeling, text classification and text matching.
Below we first introduce the unique modules, followed by the traditional tasks.

\subsection{Key Modules}
Since the implementation of fine-grained NER depends on semantic expansion, we first present semantic expansion, then fine-grained NER, and finally deep semantic representation. 

\subsubsection{Semantic Expansion}
Given an entity within a sentence, the semantic expansion module suggests a list of entities related to the given entity. For example in Figure~\ref{fig:key-feat}, the suggestion results for ``Captain Marvel'' include ``Spider-Man'', ``Captain America'', and other related movies.
Semantic expansion attaches additional information to an entity mention, which could be leveraged by upper-layer applications for better understanding the entity and the source sentence.
Possible applications of the expansion results include web search (e.g., for query suggestion) and recommendation systems.

\begin{figure}[htb]
	   \center{\includegraphics[width=\textwidth]
	       {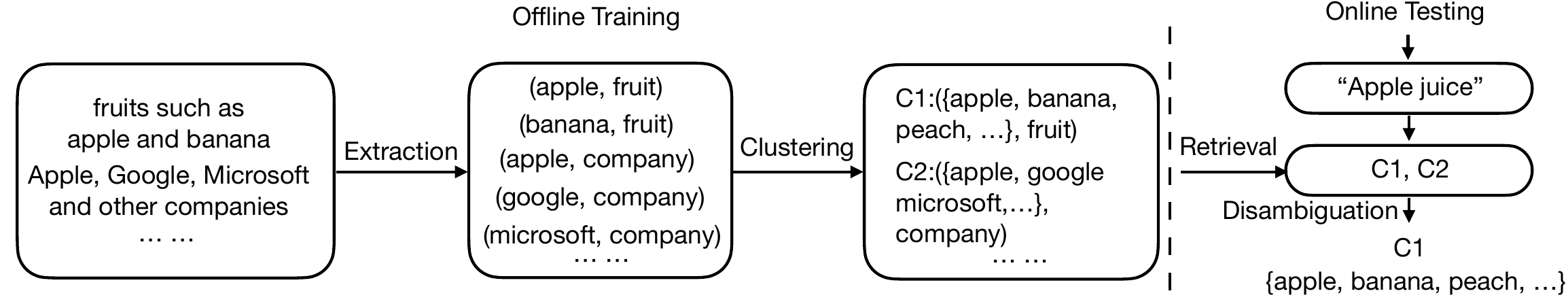}}
	  
	  \caption{\label{fig:se} Key steps for semantic expansion: extraction, clustering, retrieval and disambiguation. The first two steps are conducted offline and the last two are performed online.}
	\end{figure}

Semantic expansion task was firstly introduced in~\cite{han2020case}, and it was addressed by a neural method. However, this method is not as efficient as one expected for some industrial applications. Therefore, we propose a light-weight alternative approach in TexSmart for this task.

This approach includes two offline steps and two online ones, as illustrated in Figure~\ref{fig:se}.
During the offline procedure, Hearst patterns are first applied to a large-scale text corpus to obtain a is-a map (or called a hyponym-to-hypernym map) \cite{hearst1992automatic, zhang2011nonlinear}.
Then a clustering algorithm is employed to build a collection of term clusters from all the hyponyms, allowing a hyponym to belong to multiple clusters. Each term cluster is labeled by one or more hypernyms (or called type names).
Term similarity scores used in the clustering algorithm are calculated by a combination of word embedding, distributional similarity, and pattern-based methods \cite{mikolov2013distributed, song2018directional, shi2010corpus}.

During the online testing time, clusters containing the target entity mention are first retrieved by referring to the cluster collection. Generally, there may be multiple (ambiguous) clusters containing the target entity mention and thus it is necessary to pick the best cluster through disambiguation.
Once the best cluster is chosen, its members (or instances) can be returned as the expansion results.

Now the core challenge is how to calculate the score of a cluster given an entity mention.
We choose to compute the score as the average similarity score between a term in the cluster and a term in the context of the entity mention.
Formally, suppose $e$ is a mention in a sentence, context $\mathbf{C}=\{c_1, c_2, \cdots, c_m\}$ is a window of $e$ within the sentence, and $\mathbf{L}=\{e_1, e_2, \cdots, e_n\}$ is a term cluster containing the entity mention (i.e., $e \in \mathbf{L}$). The cluster score is then calculated below:
\begin{equation}
  \textrm{sim}(\mathbf{C}, \mathbf{L}; e) = \frac{1}{(m-1)\times (n-1)}\sum_{x\in \mathbf{C}\setminus \{e\}, y\in \mathbf{L}\setminus\{e\}} \cos(v_x, w_y)
  \label{eq:sim}
\end{equation}
\noindent where $\mathbf{C} \setminus \{e\}$ means excluding a subset $\{e\}$ from a set $\mathbf{C}$, $v_x$ denotes the input word embedding of $x$, $w_y$ denotes the output word embedding of $y$ from a well-trained word embedding model, and $\cos$ is the cosine similarity function.

It is worth mentioning that although it takes considerable time for clustering hyponyms during offline time because there are millions of hyponyms, this procedure can be conducted offline for one single time.
The online steps can be performed very efficiently.

\subsubsection{Fine-Grained NER}
\begin{figure}[htb]
  \center{\includegraphics[width=\textwidth]
      {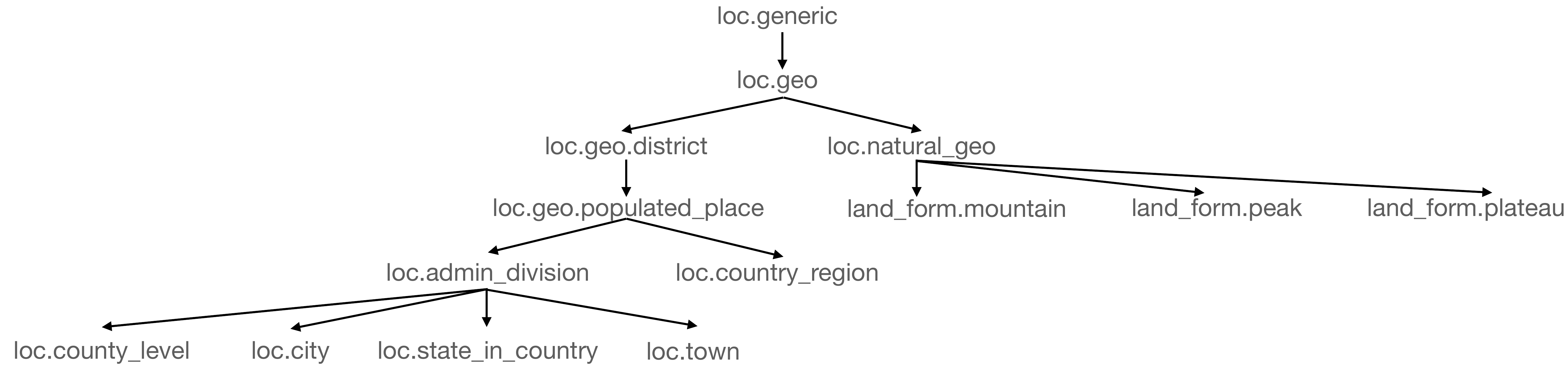}}
 \caption{\label{fig:onto} A sub-tree of the TexSmart ontology, with ``loc.generic'' as the root}
\end{figure}

Generally, it is challenging to build a fine-grained NER system. Ref \cite{xu2020clue} creates a fine-grained NER dataset for Chinese, but the number of its types is less than 20. A knowledge base (such as Freebase~\cite{bollacker2008freebase}) is utilized in \cite{ling2012fine} as distant supervision to obtain a training dataset for fine-grained NER. However, this dataset only includes about one hundred types whereas TexSmart supports up to one thousand types. Moreover, the fine-grained NER module in TexSmart does not rely on any knowledge bases and thus can be readily extended to other languages for which there is no knowledge base available. 

\paragraph{Ontology} To establish fine-grained NER in TexSmart, we need to define an ontology of entity types.
The TexSmart ontology was built in a semi-automatic way, based on the term clusters in Figure~\ref{fig:se}.
Please note that each term cluster is labeled by one or more hypernyms as type names of the cluster.
We first conduct a simple statistics over the term clusters to get a list of popular type names (i.e., those having a lot of corresponding term clusters).
Then we manually create one or more \textit{formal types} from one popular type name and add the type name to the name list of the formal types.
For example, formal type ``work.movie'' is manually built from type name ``movie'', and the word ``movie'' is added to the name list of ``work.movie''.
As another example, formal types ``language.human\_lang'' and ``language.programming'' are manually built from type name ``language'', and the word ``language'' is added to the name lists of both the two formal types.
Each formal type is also assigned with a \textit{sample instance list} in addition to a name list. Instances can be chosen manually from the clusters corresponding to the names of the formal type.
To reduce manual efforts, the sample instance list for every type is often quite short.
The supertype/subtype relation between the formal types are also specified manually.
As a result, we obtain a type hierarchy containing about 1,000 formal types, each assigned with a standard id (e.g., work.movie), a list of names (e.g., ``movie'' and ``film''), and a short list of example instances (e.g., ``Star Wars'').
The TexSmart ontology is available on the download page\footnote{\url{https://ai.tencent.com/ailab/nlp/texsmart/en/download.html}}.
Figure~\ref{fig:onto} shows a sub-tree (with type id ``loc.generic'' as the root) sampled from the entire ontology.

\paragraph{Unsupervised method}
The unsupervised fine-grained NER method works in two steps. First, run the semantic expansion algorithm (referring to the previous subsection) to get the best cluster for the entity mention. Second, derive an entity type from the cluster.

For the best cluster obtained in the first step, it contains a list of terms as instances and is also labeled with a list of hypernyms (or type names).
The final entity type id for the cluster is determined by a type scoring algorithm.
The candidate types are those in the TexSmart ontology whose name lists contain at least one hypernym of the cluster.
Please note that each entity type in the TexSmart ontology has been assigned with a name list and a sample instance list.
Therefore the score of a candidate entity type can be calculated according to the information of the entity type and cluster.

This unsupervised method has a major drawback: It cannot recognize unknown entity mentions (i.e., entity mentions that are not in any of our term clusters).

\paragraph{Hybrid method}

In order to address the above issue, we propose a hybrid method for fine-grained NER. Its key idea is to combine the results of the unsupervised method and those of a coarse-grained NER model. We train a coarse-grained NER model in a supervised manner using an off-the-shelf training dataset (for example, Ontonotes dataset~\cite{weischedel2013ontonotes}). Given the supervised and unsupervised results, the combination policy is as follows:
If the fine-grained type is compatible with the coarse type, the fine-grained type is returned; otherwise the coarse type is chosen.
Two types are compatible if one is the subtype of the other.

For example, assume that the entity mention ``apple'' in the sentence ``...apple juice...'' is determined as ``food.fruit'' by the unsupervised method and ``food.generic'' by the supervised model.
The hybrid approach returns ``food.fruit'' according to the above policy.
However, if the unsupervised method returns ``org.company'', the hybrid approach will return ``food.generic'' because the two types returned by the supervised method and the unsupervised method are not compatible.

\subsubsection{Deep Semantic Representation}
For a time or quantity entity within a sentence, TexSmart can analyze its potential structured representation, so as to further derive its precise semantic meaning. 
For example in Figure~\ref{fig:key-feat}, the deep semantic representation given by TexSmart for ``22 months ago'' is a structured string with a precise date in JSON format: \{"value": [2019, 2]\}.
Deep semantic representation is important for applications like task-oriented chatbots, where the precise meanings of some entities are required.
So far, most public text understanding tools do not provide such a feature. As a result, applications using these tools have to implement deep semantic representation by themselves.

Some NLP toolkits make use of regular expressions or supervised sequence tagging methods to recognize time and quantity entities. However, it is difficult for those methods to derive structured or deep semantic information of entities. To overcome this problem, time and quantity entities are parsed in TexSmart by Context Free Grammar (CFG), which is more expressive than regular expressions. Its key idea is similar to that in \cite{shi2015automatically} and can be described as follows: First, CFG grammar rules are manually written according to possible natural language expressions of a specific entity type. Second, the Earley algorithm~\cite{earley1970efficient} is employed to parse a piece of text to obtain semantic trees of entities. Finally, deep semantic representations of entities are derived from the semantic trees.

\subsection{Other Modules}

\subsubsection{Word Segmentation}
\begin{figure}
	\centering  
	\includegraphics[scale=0.58]{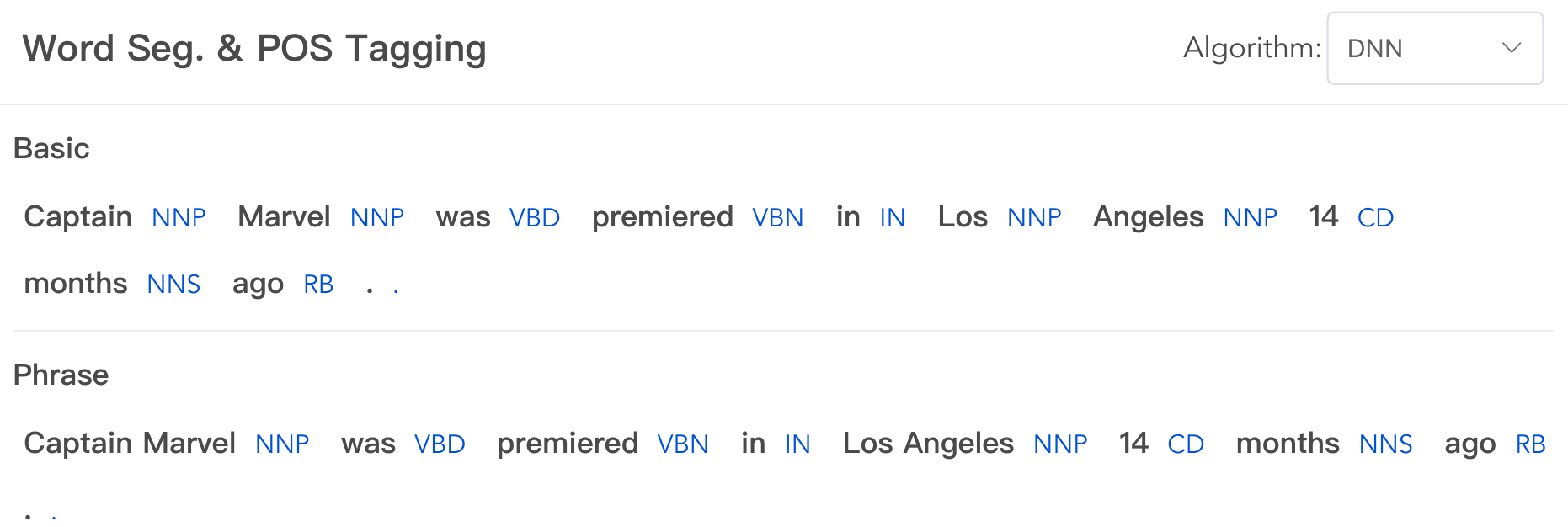}  
	\caption{An example of English word segmentation and part-of-speech tagging.}  
	\label{word_seg_en}  
\end{figure}

\begin{figure}
	\centering  
	\includegraphics[scale=0.58]{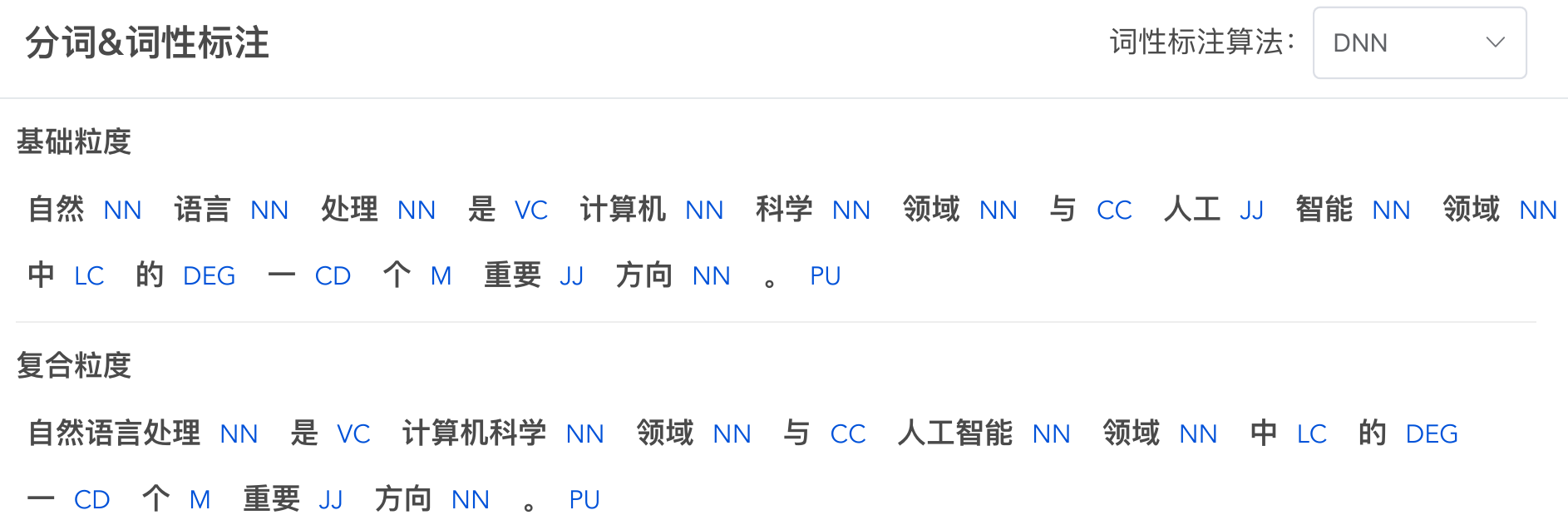}  
	\caption{An example of Chinese word segmentation and part-of-speech tagging.}
	\label{word_seg_zh}  
\end{figure}

Automatic word segmentation is a task for segmenting a piece of text into a word sequence. 
The results of word segmentation are widely used in many downstream NLP tasks.
In order to support different application scenarios, TexSmart provides word segmentation results of two granularity levels: word level (or basic level), and phrase level.
Figure \ref{word_seg_en} shows an example of English word segmentation results.
As shown in the figure, some phrases (especially noun phrases) are contained in the phrase-level segmentation results.
An example of Chinese word segmentation results is shown in Figure \ref{word_seg_zh}.

An unsupervised algorithm is implemented in TexSmart for both English and Chinese word segmentation.
We choose an unsupervised method over supervised ones due to two reasons. First, it is at least 10 times faster. Second, its accuracy is good enough for most applications.

\subsubsection{Part-of-Speech Tagging}
Part-of-Speech (POS) denotes the syntactic role of each word in a sentence, also known as word classes or syntactic categories.
Popular part-of-speech tags include noun, verb, adjective, etc.
The task of POS Tagging is about assigning a part-of-speech tag to each word in the input text.
POS tagging information is important in many information extraction tasks.
For some text understanding tasks like NER, the performance of some algorithms can be improved if POS tagging information is included.
Part-of-speech tags also help us to construct good sentences.

In Figure \ref{word_seg_en} and \ref{word_seg_zh},  in addition to word segmentation results, we also present the POS tag of each word in the sentence under different word segmentation granularities.
The Chinese and English part-of-speech tag sets supported by TexSmart are CTB\footnote{\url{https://ai.tencent.com/ailab/nlp/texsmart/table\_html/table2.html}}  \cite{xue2005penn} and PTB\footnote{\url{https://ai.tencent.com/ailab/nlp/texsmart/table\_html/table3.html}} \cite{Santorini1990PartofSpeechTG},  respectively.

We implement three models for part-of-speech tagging: Log-linear based, conditional random field (CRF) based and deep neural network (DNN) based models~\cite{ratnaparkhi1996maximum,lafferty2001conditional,akbik2018coling,liu2019roberta}.
We denote them as: log\_linear, crf and dnn, respectively.
 In the web version of the demo, users can select different algorithms in the drop-down box, i.e., ``词性标注算法'' in the Chinese page (Figure \ref{word_seg_zh}) or ``Algorithm'' in the English page (Figure \ref{word_seg_en}).
 When the HTTP API is called, one of the three algorithms can be chosen freely.

\subsubsection{Coarse-grained NER}
Coarse-grained NER  aims to identify entity mentions from a sentence and assigns them with coarse-grained entity types, such as Person, Location, Organization, etc, rather than the fine-grained types as described in Section 2.1.1 for fine-grained NER.
For example, given the following sentence:

\emph{``Captain Marvel was premiered in Los Angeles 22 months ago.''},

the mention ``Los Angeles'' is identified by the coarse-grained NER algorithm with the coarse type ``loc.generic''.

The difference between fine-grained and coarse-grained NERs is that the former involves more entity types with a finer granularity.
Since coarse-grained NER includes much fewer entity types,\footnote{The type set considered in TexSmart and its descriptions can be found in \url{https://ai.tencent.com/ailab/nlp/texsmart/table\_html/table4.html}.} it has its own advantages: (1) with a faster speed,  (2) having a higher precision, which makes it suitable for some special applications.

We implement coarse-grained NER using supervised learning methods, including conditional random field (CRF)~\cite{lafferty2001conditional} based and deep neural network (DNN) based models~\cite{akbik2018coling,liu2019roberta,li2020segmenting}.

\begin{figure}
	\centering  
	\includegraphics[scale=0.58]{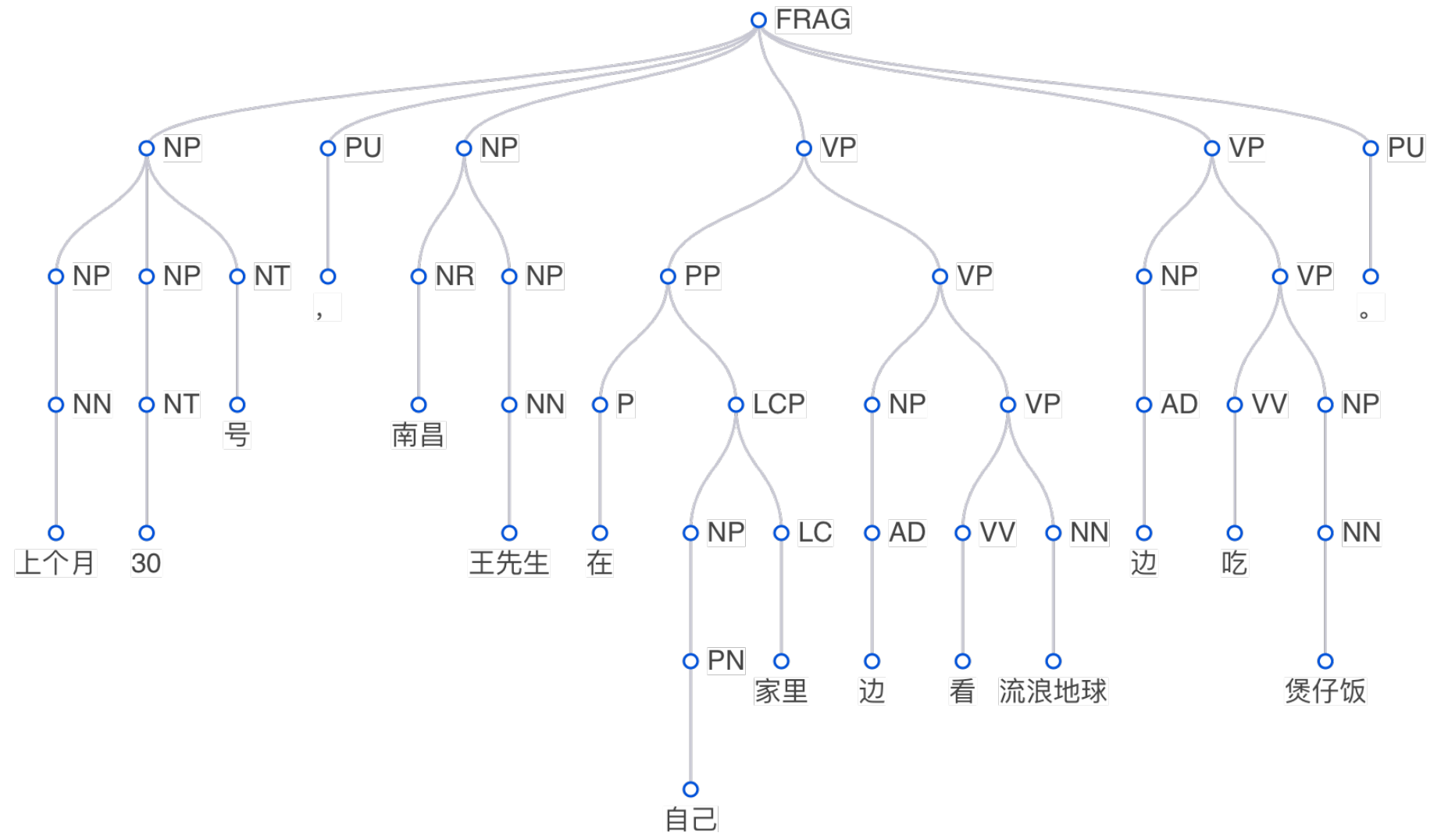}  
	\caption{An example of the constituency parsing task.}  
	\label{Constituency_Parsing}  
\end{figure}

\subsubsection{Constituency Parsing}
Constituency parsing is used to parse an input word sequence into the corresponding phrase structure tree, which is based on the phrase structure grammar proposed by Chomsky.
The phrase structure includes noun phrases (NP), verb phrases (VP), etc.
For example, in Figure \ref{Constituency_Parsing},  ``南昌 (Nanchang)'' or ``王先生 (Mr Wang)'' constitutes a noun phrase, and ``吃 (eat)'' or ``煲仔饭 (claypot rice)'' constitutes a verb phrase.

We present an example to illustrate constituency parsing.
The leaf node in the phrase structure tree, such as ``流浪地球 (The Wandering Earth)''  in  Figure \ref{Constituency_Parsing}, denotes a word of the input sentence.
The  parent node (e.g., ``NN'') of a leaf node is the part-of-speech tag of the word corresponding to the leaf node.
Further, the parent node (e.g.,  ``VP'') of the part-of-speech tag (e.g, ``NN'')  node refers to the part-of-speech tag of the phrase structure that constitutes the leaf nodes, e.g., ``看 (watch)'' and ``流浪地球 (The Wandering Earth)''  in  Figure \ref{Constituency_Parsing}.  
The phrase structure labels used in TexSmart can be found in the home-page of TexSmart.\footnote{ \url{https://ai.tencent.com/ailab/nlp/texsmart/table\_html/table6.html} (CTB, Chinese) and \url{https://ai.tencent.com/ailab/nlp/texsmart/table\_html/table7.html} (PTB, English), respectively.}

We implement the constituency parsing model based on the work \cite{Kitaev2018ConstituencyPW}. 
 Ref \cite{Kitaev2018ConstituencyPW}  builds the parser by combining a  sentence encoder with a chart decoder based on the self-attention mechanism.
 Different from work \cite{Kitaev2018ConstituencyPW} , we use pre-trained BERT model as the text encoder to extract features to support the subsequent decoder-based parsing.
Our model achieves excellent performance and has low search complexity.

\begin{figure}
	\centering  
	\includegraphics[scale=0.68]{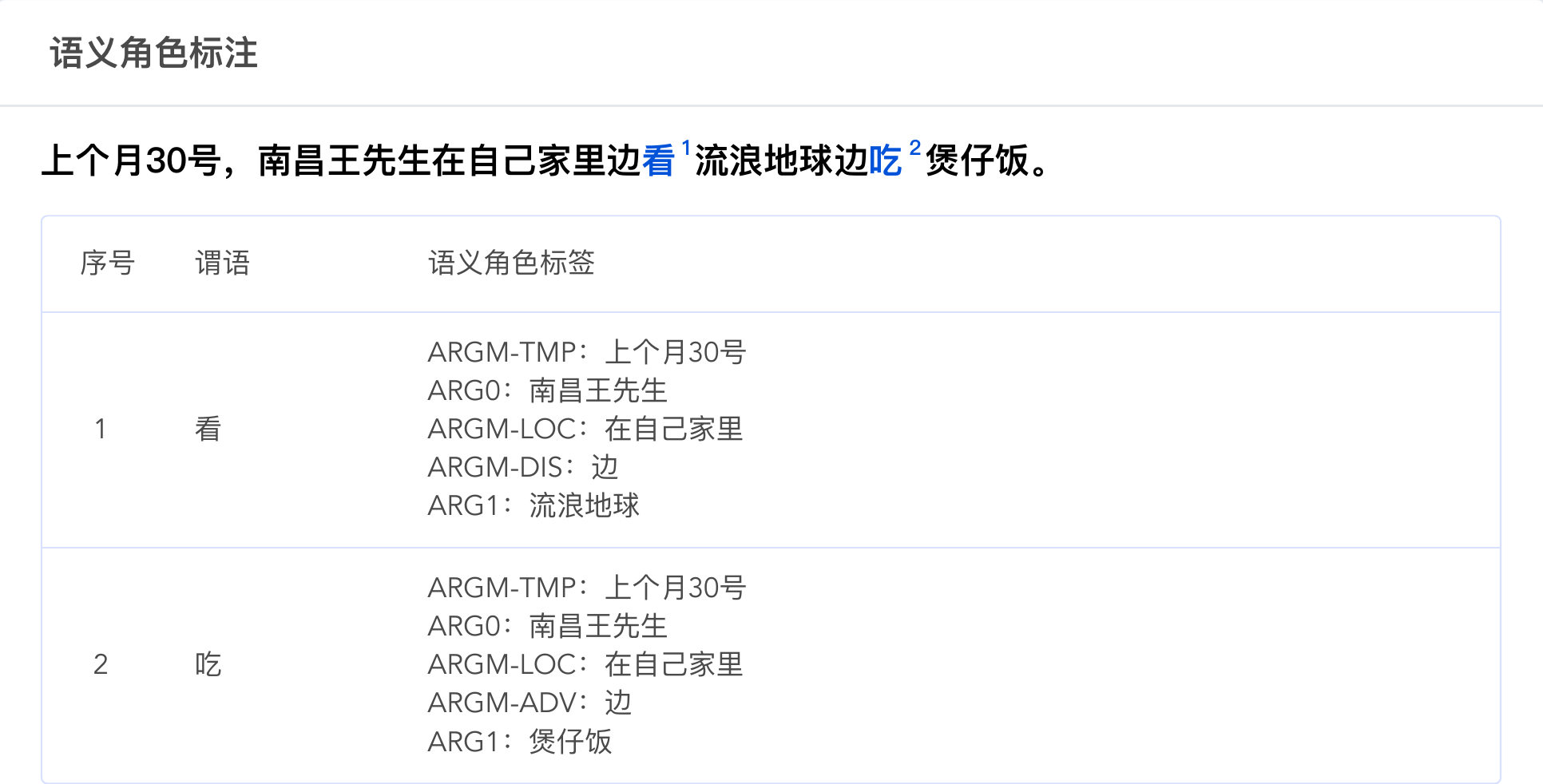}  
	\caption{An example to illustrate the task of semantic role labeling.}  
	\label{srl}  
\end{figure}

\subsubsection{Semantic Role Labeling}
In natural language processing, semantic role labeling (also called shallow semantic parsing) tries to assign role labels to words or phrases in a sentence.
The labels indicate the semantic role of words in the sentence,  including the basic elements of a event: trigger word, subject, object and other adjuncts, such as time, location, method, etc. of the event.
The task of semantic role labeling is closely related to event semantics, where the elements that make up an event are called semantic roles.
In semantic role labeling, we need to identify all the events and their basic elements in a sentence.

We present an example to illustrate semantic role labeling.
In Figure \ref{srl},  two events with trigger words of  ``看 (watch)''  and ``吃 (eat)'' are detected by TexSmart.
Further, all the elements of  the two events are also detected.
For example, given  the trigger word ``看 (watch)'',  the subject (ARG0),  object (ARG1),  adjunct time (ARGM-TMP) and adjunct location (ARGM-LOC) are ``南昌的王先生 (Mr. Wang in Nanchang)'', ``流浪地球 (The Wandering Earth)'',  ``上个月30号(On the 30th of last month)'' and ``在自己家里 (in his own home)'', respectively.

TexSmart takes a sequence labeling model with BERT as the text encoder for semantic role labeling similar to~\cite{shi2019simple}.
The results of semantic role labeling support many downstream tasks, such as deeper semantic analysis (AMR Parsing, CCG Parsing, etc.), intention recognition in a task-oriented dialogue system, entity scoring in knowledge-based question answering, etc.
TexSmart supports semantic role labeling on both Chinese and English texts.
The Chinese label set used can be found in the home-page of TexSmart.\footnote{\url{ https://ai.tencent.com/ailab/nlp/texsmart/table\_html/table8.html}, and the English are presented in \url{https://ai.tencent.com/ailab/nlp/texsmart/table\_html/table9.html}.}

\subsubsection{Text Classification}
Text Classification aims to assign a semantic label for an input text among a predefined label set. 
Text Classification is a classical task in NLP and it has been widely used in many applications, such as spam filtering,  sentiment analysis and question classification. 
For example, given the input text:

``\emph{TensorFlow (released as open source on November 2015) is a library developed by Google to accelerate deep learning research.}'', 

TexSmart classifies it into a category named ``tc.technology''  with a confidence score of 0.198.
The predefined label set in TexSmart is available on the web page.\footnote{ \url{https://ai.tencent.com/ailab/nlp/texsmart/table\_html/tc\_label\_set.html}.}

\subsubsection{Text Matching}
Text matching is the task of calculating the semantic similarity between two pieces of text.
It has wide applications in NLP, information retrieval, Web search, and recommendation systems.
For example, it is a core task in information retrieval to estimate the similarity between an input query and a document.
As another example, question answering is usually modeled as the matching between a question and its answer candidate.

We implement two text matching algorithms in TexSmart: Linkage and ESIM \cite{Chen2017EnhancedLF}.
Linkage is an unsupervised algorithm designed by ourselves that incorporates synonymy information and word embedding knowledge to compute semantic similarity.
ESIM is the abbreviation of ``Enhanced LSTM for Natural Language Inference''.
Different from the previous models with complicated network architectures, ESIM carefully designs the sequential 
model with both local and global inference based on chain LSTMs and outperforms the counterparts.

For training the ESIM model for Chinese, we automatically created a large-scale labeled dataset of Chinese sentences, to ensure a better generalization capacity than the models trained from small datasets.
Since the English training data is not ready yet, the English version of ESIM is not available so far.
We would like to leave it as future work.

\section{System Usage}
Two ways are available to use TexSmart: Calling the HTTP API directly, or downloading one version of the offline SDK.
Note that for the same input text, the results from the HTTP API and the SDK may be slightly different, because the HTTP API employs a larger knowledge base and supports more text understanding tasks and algorithms.
The detailed comparison between the SDK and the HTTP API is available in

\url{https://ai.tencent.com/ailab/nlp/texsmart/en/instructions.html}.

The latest and history versions of the SDK can be downloaded from  

\url{https://ai.tencent.com/ailab/nlp/texsmart/en/download.html},

and detailed information about how to call the TexSmart APIs is introduced in 

\url{https://ai.tencent.com/ailab/nlp/texsmart/en/api.html}.

\subsection{Offline Toolkit (SDK)}

So far the SDK supports Linux, Windows, and Windows Subsystem for Linux (WSL). Mac OS support will be added in v0.3.0.
Programming languages supported include C, C++, Python (both version 2 and version 3) and Java (version $\geq$ 1.6.0).

Example codes for using the SDK with different programming languages are in the ./examples sub-folder.
For example, the Python codes in ./examples/python/en\_nlu\_example1.py show how to use the TexSmart SDK to process an English sentence.
The C++ codes in ./examples/c\_cpp/src/nlu\_cpp\_example1.cc show how to use the SDK to analyze both an English sentence and a Chinese sentence.

\subsection{TexSmart HTTP API}
The HTTP API of TexSmart contains two parts: the text understanding API and the text matching API.

\subsubsection{Text Understanding API}
The text understanding API can be accessed via HTTP-POST and the URL is available on the web page\footnote{https://texsmart.qq.com/api}.

Both the input post-data and the response are in JSON format.
It is recommended to use Postman\footnote{\url{https://www.postman.com/downloads/}} to test the API.
Below is a sample input:

\emph{\{"str": "He stayed in San Francisco."\}}

\begin{figure}
	\centering  
	\includegraphics[scale=0.82]{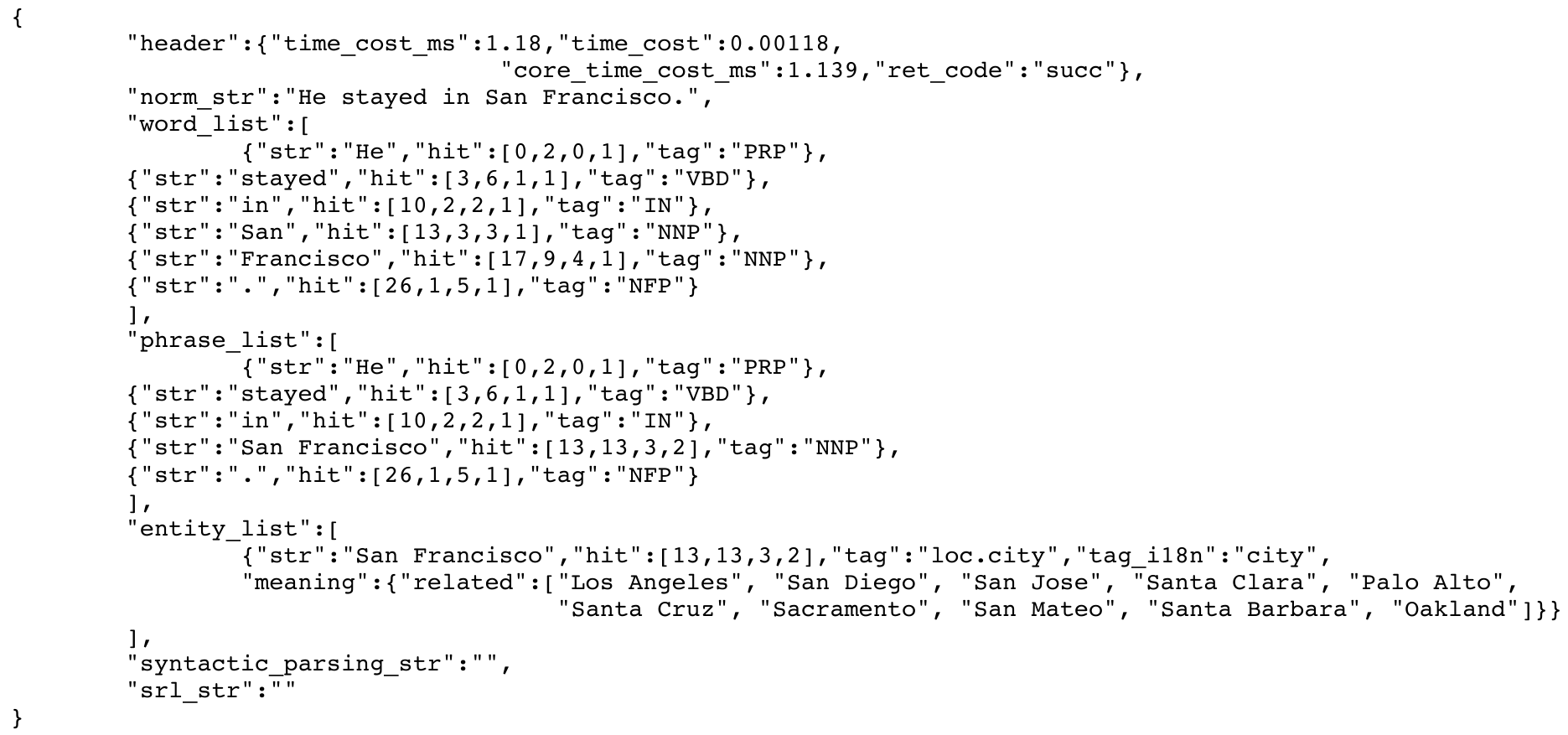}  
	\caption{The JSON results of the text understanding API for input sentence ``He stayed in San Francisco.''} . 
	\label{tex_under_http_api}  
\end{figure}

The results are shown in Figure \ref{tex_under_http_api}.
Note that, in this example, both fields ``syntactic\_parsing\_str'' and ``srl\_str'' have empty strings, because syntactic parsing and semantic role labeling are not enabled by default.
We summarize the fields used and their descriptions in Table \ref{fields_and_descriptions_text_understanding_API}.
More details about the output JSON contents is available on the web page.\footnote{\url{https://ai.tencent.com/ailab/nlp/texsmart/en/api.html}.}

 \begin{table}[!htbp]
 	\centering
 	\caption{Major fields of the output JSON object returned by the text understanding API}
 	\begin{tabular}{|p{3.3cm}|p{12.3cm}|}
 		\hline
 		Field  & Description \\
 		\hline
 		header	& Containing some auxiliary information about this API call, including time cost, error codes, etc \\
 		\hline
 		norm\_str&	Normalized text	\\
 		\hline
 		word\_list &Results of word-level word segmentation and POS tagging \\
 		\hline
 		phrase\_list 	&Results of phrase-level word segmentation and POS tagging \\
 		\hline
 		entity\_list& Results of NER, semantic expansion, and deep semantic representation \\
 		\hline
 	    syntactic\_parsing\_str& Results of constituency syntax parsing  \\
 	\hline
 	    srl\_str	& Results of semantic role labeling  \\
 		\hline
 		cat\_list & Results of text classification \\
 		\hline
	 \end{tabular}
 	\label{fields_and_descriptions_text_understanding_API}
 \end{table}

\begin{figure}
	\centering  
	\includegraphics[scale=0.70]{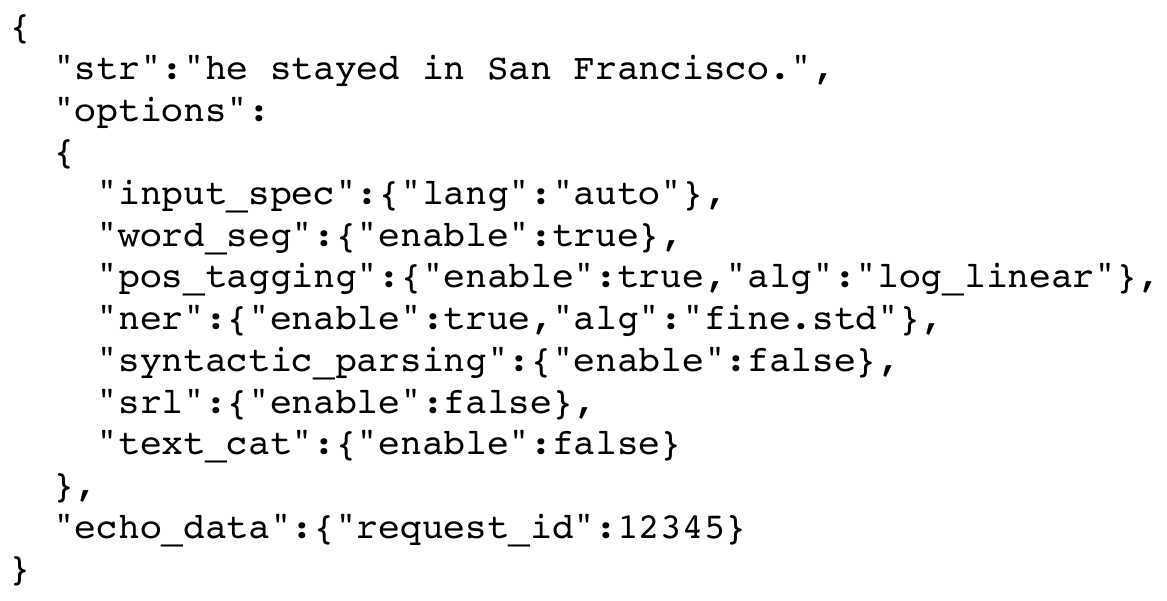}  
	\caption{An example input JSON containing options}  
	\label{defalt_json}  
\end{figure}

\paragraph{Advanced options}
Some options can be added to enable/disable some modules, and to specify the algorithms used in some modules.
Some key options are summarized in Table~\ref{fields_and_descriptions_text_understanding_input_API}.
Figure~\ref{defalt_json} shows an example input JSON with options.

 \begin{table}[!htbp]
	\centering
	\caption{Major fields of the options JSON object in the text understanding API. Field x.y means field y in the value of field x.}
	\begin{tabular}{|p{4cm}|p{3.6cm}|p{7.6cm}|}
		\hline
		Field  & Values & Description \\
		\hline
		input\_spec.lang& \{auto, chs,  en\}	  &Specify the language of the input text. The default value is ``auto'', which means the language is detected automatically.\\
		\hline
		word\_seg.enable& \{true, false\} 	&Enable or disable the word segmentation module.\\
		\hline
		pos\_tagging.enable& \{true, false\} 	&Enable or disable the POS tagging module.\\
		\hline
		pos\_tagging.alg& \{log\_linear, crf, dnn\} 	&The algorithm used in the POS tagging module.  \\
		\hline
		ner.enable& \{true, false\} 	&Enable or disable the NER module.\\
		\hline
	ner.alg& \{coarse.crf, coarse.dnn, coarse.lua, fine.std, fine.high\_acc\}	&The algorithm used in the NER module.\\
	    \hline
		syntactic\_parsing.enable& \{true, false\} 	&Enable or disable the syntactic parsing module.\\
		\hline
		srl.enable& \{true, false\} 	&Enable or disable the semantic role labeling module.\\
		\hline
	\end{tabular}
	\label{fields_and_descriptions_text_understanding_input_API}
\end{table}

\paragraph{Batch mode}

\begin{figure}
	\centering  
	\includegraphics[scale=0.85]{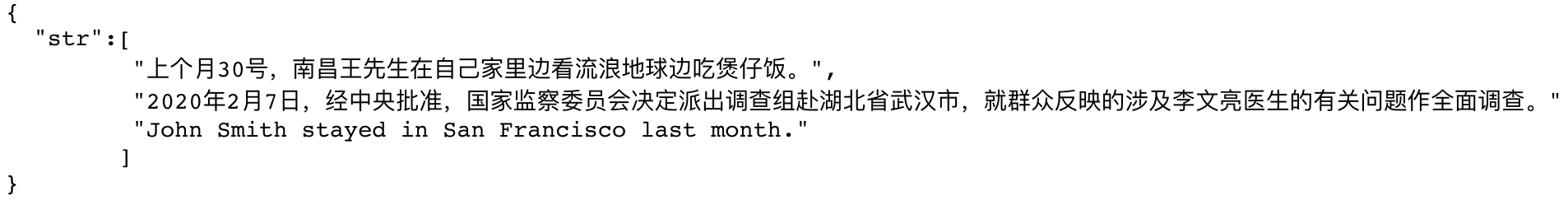}  
	\caption{An example to illustrate batch call with multiple languages in text understanding API.}  
	\label{Batch_Call}  
\end{figure}

\begin{figure}
	\centering  
	\includegraphics[scale=0.75]{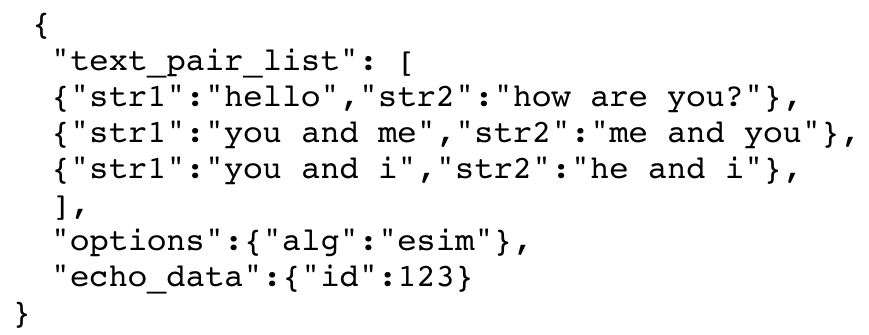}  
	\caption{An example input of text matching API.}  
	\label{tex_match_http_api}  
\end{figure}

\begin{figure}
	\centering  
	\includegraphics[scale=0.75]{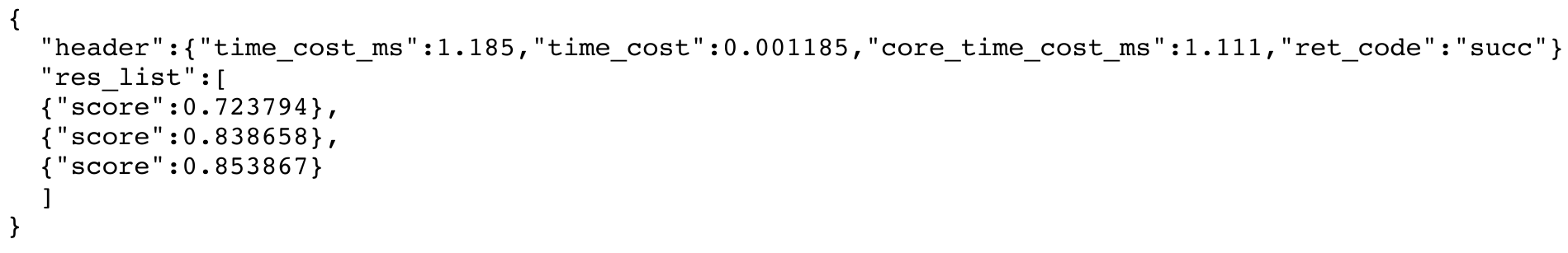}  
	\caption{The output of the text matching API for the input text in Figure \ref{tex_match_http_api}.}  
	\label{tex_match_output_api}  
\end{figure}

Multiple sentences (of possibly different languages) can be processed in one API call.
An example is shown in Figure \ref{Batch_Call}, where one English sentence and two Chinese sentences are processed in one API call.
Note that the output format of the batch mode is slightly different from that of the ordinary mode. In the batch mode, the results for the input sentences form a JSON array, as the value of the ``res\_list'' field.

\subsubsection{Text Matching API}
The text matching API is used to calculate the similarity between a pair of sentences.
Similar to the text understanding API, the text matching API also supports access via HTTP-POST and the URL is available on the web page.\footnote{
\url{https://texsmart.qq.com/api/match\_text}.}

Figure \ref{tex_match_http_api} shows an example input JSON. The corresponding response returned by the API is shown in Figure \ref{tex_match_output_api}.

\section{System Evaluation}

\subsection{Settings}

\paragraph{Semantic Expansion}
The performance of semantic expansion are evaluated based on human annotation.
We first select at random 5,000 \textit{<sentence, entity mention>} pairs (called SE pairs) from our test set of NER (to make sure that the entities selected are correct).
Then our semantic expansion algorithm is applied to the SE pairs to generate a related-entity list for each pair.
Top nine expansion results of each SE pair are then judged by human annotators in terms of quality and relatedness, with each result annotated by two annotators.
For each result, a label of 2, 1, or 0 is assigned by each annotator. The three labels mean ``highly related'', ``slightly related'', and ``not related'' respectively.
In calculating evaluation scores, the three labels are normalized to scores 100, 50, and 0 respectively.
The score of each result is the average of the scores given by the two annotators.

\paragraph{Fine-grained NER}
Ref ~\cite{ling2012fine} provides a test set for fine-grained NER evaluation. However, this dataset only contains about 400 sentences. In addition, it misses some important entities during human annotation, which is a common issue in building a dataset for evaluating fine-grained NER~\cite{li2020empirical}.
Therefore, we create a larger fine-grained NER dataset, based on the Ontonotes 5.0 dataset.
We ask human annotators to label fine-grained types for each coarse-labeled entity. Since human annotators do not need to identify mentions from scratch, it would mitigate the missing entities issue to some extent.
Furthermore, because it is too costly for human annotators to annotate types from the entire ontology, we instead take a sub-ontology for human annotation which combines all types from~\cite{ling2012fine} and~\cite{gillick2014context}, including 140 types in total. 

To set the hybrid method for fine-grained NER, we select LUA~\cite{li2020segmenting} as the coarse-grained NER model, which is trained on Ontonotes 5.0 training dataset~\cite{weischedel2013ontonotes}. 
To compare fine-grained NER against coarse-grained NER, we report a variant of F1 measure for evaluation which only differs from standard F1 in matching count accumulation: if an output type is a fine-grained type and it exactly matches a gold fine-grained type, the matching count accumulates 1; if an output is a coarse grained type and it is compatible with a gold fine-grained type, the matching count accumulates 0.5.

\paragraph{POS Tagging}
We evaluate three POS tagging algorithms: log-linear, CRF, and DNN. They are all trained on PTB for English and CTB 9.0 for Chinese.
We use their corresponding test sets to evaluate all the models.

\paragraph{Coarse-grained NER}
To ensure better generalization to industrial applications, we combine several public training sets together for English NER.
They are CoNLL2003~\cite{sang2003introduction}, BTC~\cite{derczynski2016broad}, GMB~\cite{Bos2017GMB}, SEC\_FILING~\cite{alvarado2015domain}, WikiGold~\cite{balasuriya2009named,nothman2013learning}, and WNUT17~\cite{derczynski2017results}. 
Since the label set for all these datasets are slightly different, we only maintain three common labels (Person, Location and Organization) for training and testing.
For Chinese, we create a NER dataset including about 80 thousand sentences labeled with 12 entity types, by following a similar guideline to that of the Ontonotes dataset. We randomly split it into a training set and a test set with ratio of 3:1. 
We evaluate two algorithms for coarse-grained NER: CRF and DNN. For DNN, we implement the RoBERTa-CRF and Flair models. As we found RoBERTa-CRF performs better on the Chinese dataset while Flair is better on the English dataset, we report results of RoBERTa-CRF for Chinese and Flair for English in our experiments.

\begin{table}[htb]
	\centering
	\caption{\label{tab:se+fgner} Semantic expansion and fine-grained NER evaluation results. Semantic expansion is evaluated by the averaged score from human annotators. Fine-grained NER is evaluated by a variant of F1 score which can measure both coarse-grained and fine-grained models against a fine-grained ground-truth dataset.}
	\begin{tabular}{c|c|c|c|c}
	\hline\hline
	\multirow{2}{*}{} & \multicolumn{2}{c|}{Semantic Expansion (human score)} & \multicolumn{2}{c}{Fine-grained NER (F1)} \\ \cline{2-5} 
								& ZH  & EN  & LUA                 & Hybrid               \\ \hline
	Quality (human score or F1) &   79.5                      & 80.5                      & 45.9                & 53.8                 \\ \hline\hline
	\end{tabular}
\end{table}

\begin{table}[htb]
  \centering
\caption{\label{tab:pos+ner} Evaluation results for some POS Tagging and coarse-grained NER algorithms in TexSmart on both English (EN) and Chinese (ZH) datasets. The English and Chinese NER datasets are labeled with 3 and 12 entity types respectively.}
  \begin{tabular}{c|c|c|c|c|c|c|c|c|c|c}
  \hline\hline
  \multirow{3}{*}{} & \multicolumn{6}{c|}{POS Tagging}                                                       & \multicolumn{4}{c}{Coarse-grained NER}                  \\ \cline{2-11} 
                    & \multicolumn{2}{c|}{Log-linear} & \multicolumn{2}{c|}{CRF}  & \multicolumn{2}{c|}{DNN} & \multicolumn{2}{c|}{CRF} & \multicolumn{2}{c}{DNN}      \\ \cline{2-11} 
                    
                    & EN             & ZH             & EN          & ZH          & EN          & ZH         & EN          & ZH         & EN             & ZH           \\ \hline
  F1                & 96.76          & 93.94          & 96.50        & 93.73       &  97.04           &     98.08       &       73.24      &   67.26         & 83.12          & 75.23        \\ \hline
  Sents/sec         & \multicolumn{2}{c|}{3.9K}       & \multicolumn{2}{c|}{1.3K} & \multicolumn{2}{c|}{149}    & \multicolumn{2}{c|}{1.1K}    & \multicolumn{2}{c}{107} \\ \hline\hline
  \end{tabular}

  \end{table}

\paragraph{Constituency Parsing}
We conduct parsing experiments on both English and Chinese datasets. For English task, we use WSJ sections in Penn Treebank (PTB) ~\cite{marcus1993building}, and we follow the standard splits: the training data ranges from section 2 to section 21; the development data is section 24; and the test data is section 23. For Chinese task, we use the Penn Chinese Treebank (CTB) of the version 5.1 ~\cite{xue2005penn}. The training data includes the articles 001-270 and articles 440-1151; the development data is the articles 301- 325; and the test data is the articles 271-300.

\paragraph{SRL}
Semantic role labeling experiments are conducted on both English and Chinese datasets.
We use the CoNLL 2012 datasets \cite{pradhan2013towards} and follow the standard splits for the training, development and test sets.
The network parameters of our model are initialized using RoBERTa.
The batch size is set to 32 and the learning rate is 5$\times$10$^{-5}$.

\paragraph{Text Matching}
Two text matching algorithms are evaluated: ESIM and Linkage.
The datasets used in evaluating English text matching are MRPC\footnote{\url{https://www.microsoft.com/en-us/download/details.aspx?id=52398}.} and QUORA\footnote{\url{https://www.quora.com/q/quoradata/First-Quora-Dataset-Release-Question-Pairs}.}.
For Chinese text matching, four datasets are involved: LCQMC~\cite{liu-etal-2018-lcqmc}, AFQMC~\cite{xu2020clue}, BQ\_CORPUS~\cite{chen-etal-2018-bq}, and PAWS-zh~\cite{zhang+:2018:paws}.
We evaluate the quality and speed for both ESIM and Linkage algorithms in terms of F1 score and sentences per second, respectively. 
Since we have not trained the English version of ESIM yet, the corresponding evaluation results are not reported.

  \begin{table}[h]
    \centering
    \caption{\label{tab:cp+srl} Evaluation results for constituency parsing and SRL}

    \begin{tabular}{c|c|c|c|c}
    \hline\hline
    & \multicolumn{2}{c|}{Constituency Parsing} & \multicolumn{2}{|c}{SRL} \\
    \hline
    & EN  & ZH & EN & ZH \\ \hline
    F1        &  95.42 & 92.25 & 86.7  & 82.1  \\ \hline
    Sents/sec &   16.60  &  16.00  & 10.2 & 11.5  \\ \hline\hline
    \end{tabular}
    \end{table}

\begin{table}[h]
  \centering
  \caption{\label{tab:tm} Text matching evaluation results. ESIM is a supervised algorithm and it is trained on an in-house labeled dataset only for Chinese. Linkage is an unsupervised algorithm and it is trained for both English and Chinese.}
  \begin{tabular}{c|c|c|c|c|c|c|c}
  \hline\hline
  \multirow{2}{*}{Algorithms} & \multirow{2}{*}{Sents/Sec} & \multicolumn{2}{c|}{English} & \multicolumn{4}{c}{Chinese} \\ 
                    &                   &    MRPC  & QUORA & LCQMC & AFQMC & BQ\_CORPUS & PAWS-zh \\ \hline
    ESIM  &  861    &   -  &   -  &  82.63 & 51.30 & 71.05 & 61.55 \\ \hline
    Linkage  & 1973  &  82.18   &  74.94  &  79.26 & 48.66 & 71.23 & 62.30 \\ \hline \hline
\end{tabular}
\end{table}

\subsection{Evaluation Results}

Table~\ref{tab:se+fgner} shows the evaluation results of semantic expansion and fine-grained NER. For semantic expansion, it is shown that TexSmart achieves an accuracy of about 80.0 on both English and Chinese datasets. It is a pretty good performance.
For fine-grained NER, it is observed that the hybrid approach performs much better than the supervised model (LUA).

The evaluation results for POS Tagging and coarse-grained NER are listed in Table~\ref{tab:pos+ner}.
The speed values in this table are measured in sentences per second. Please note that the speed results for Log-linear and CRF are obtained using one single thread, while the speed results for DNN are on 6 threads.

It is clear from the POS tagging results that the three algorithms form a spectrum. 
On one side of the spectrum is the log-linear algorithm, which is very fast but less accurate than the DNN algorithm. On the opposite side is the DNN algorithm, which achieves the best accuracy but are much slower than the other two algorithms. The CRF algorithm is in the middle of the spectrum.

Also from Table~\ref{tab:pos+ner}, we can see that the two coarse-grained NER algorithms form another spectrum. The CRF algorithm is on the high-speed side, while the DNN algorithm is on the high-accuracy side.
Note that for DNN methods in this table, we employ a data augmentation method to improve their generalization abilities and a knowledge distillation method to speed up its inference~\cite{hinton2015distilling}.

Evaluation results for constituency parsing and semantic role labeling are summarized in Table~\ref{tab:cp+srl}. For constituency parsing, the F1 scores on the English and Chinese test sets are 95.42 and 92.25, respectively. The decoding speed depends on the input sentence length. It can process 16.6 and 16.0 sentences per second on our test sets.
For SRL, the F1 scores on the English and Chinese test sets are 86.7 and 82.1 respectively and it processes about 10 sentences per second.
The speed may be too slow for some applications.
As future work, we plan to design more efficient syntactic parsing and SRL algorithms.

Table~\ref{tab:tm} shows the performance of two algorithms for text matching.
We can see from this table that, in terms of speed, both algorithms are fairly efficient.
Please note that the speed is measured in sentences per second using one single CPU.
In terms of accuracy, their performance comparison depends on the dataset being used. ESIM performs apparently better on the first two datasets, while slightly worse on the last one.
Applications may need to test on their datasets before making decision between the two algorithms.

\section{Conclusion}
In this technical report we have presented TexSmart, a text understanding system that supports fine-grained NER, enhanced semantic analysis, as well as some common text understanding functionalities.
We have introduced the main functions of TexSmart and key algorithms for implementing the functions.
Some instructions about how to use the TexSmart offline SDK and online APIs have been described.
We have also reported some evaluation results on major modules of TexSmart.

\bibliographystyle{unsrt}  



\bibliography{references.bib}

\begin{thebibliography}{10}

\bibitem{loper2002nltk}
Edward Loper and Steven Bird.
\newblock Nltk: the natural language toolkit.
\newblock In {\em Proceedings of the ACL-02 Workshop on Effective tools and
  methodologies for teaching natural language processing and computational
  linguistics}, 2002.

\bibitem{OpenNLP}
https://opennlp.apache.org.

\bibitem{manning2014stanford}
Christopher~D Manning, Mihai Surdeanu, John Bauer, Jenny~Rose Finkel, Steven
  Bethard, and David McClosky.
\newblock The stanford corenlp natural language processing toolkit.
\newblock In {\em Proceedings of 52nd annual meeting of the association for
  computational linguistics: system demonstrations}, pages 55--60, 2014.

\bibitem{gardner-etal-2018-allennlp}
Matt Gardner, Joel Grus, Mark Neumann, Oyvind Tafjord, Pradeep Dasigi,
  Nelson~F. Liu, Matthew Peters, Michael Schmitz, and Luke Zettlemoyer.
\newblock {A}llen{NLP}: A deep semantic natural language processing platform.
\newblock In {\em Proceedings of Workshop for {NLP} Open Source Software
  ({NLP}-{OSS})}, pages 1--6, Melbourne, Australia, July 2018. Association for
  Computational Linguistics.

\bibitem{che2010ltp}
Wanxiang Che, Zhenghua Li, and Ting Liu.
\newblock Ltp: A chinese language technology platform.
\newblock In {\em Coling 2010: Demonstrations}, pages 13--16, 2010.

\bibitem{qiu2013fudannlp}
Xipeng Qiu, Qi~Zhang, and Xuan-Jing Huang.
\newblock Fudannlp: A toolkit for chinese natural language processing.
\newblock In {\em Proceedings of the 51st Annual Meeting of the Association for
  Computational Linguistics: System Demonstrations}, pages 49--54, 2013.

\bibitem{han2020case}
Jialong Han, Aixin Sun, Haisong Zhang, Chenliang Li, and Shuming Shi.
\newblock Case: Context-aware semantic expansion.
\newblock In {\em AAAI}, pages 7871--7878, 2020.

\bibitem{hearst1992automatic}
Marti~A Hearst.
\newblock Automatic acquisition of hyponyms from large text corpora.
\newblock In {\em Coling 1992 volume 2: The 15th international conference on
  computational linguistics}, 1992.

\bibitem{zhang2011nonlinear}
Fan Zhang, Shuming Shi, Jing Liu, Shuqi Sun, and Chin-Yew Lin.
\newblock Nonlinear evidence fusion and propagation for hyponymy relation
  mining.
\newblock In {\em Proceedings of the 49th Annual Meeting of the Association for
  Computational Linguistics: Human Language Technologies}, pages 1159--1168,
  2011.

\bibitem{mikolov2013distributed}
Tomas Mikolov, Ilya Sutskever, Kai Chen, Greg~S Corrado, and Jeff Dean.
\newblock Distributed representations of words and phrases and their
  compositionality.
\newblock {\em Advances in neural information processing systems},
  26:3111--3119, 2013.

\bibitem{song2018directional}
Yan Song, Shuming Shi, Jing Li, and Haisong Zhang.
\newblock Directional skip-gram: Explicitly distinguishing left and right
  context for word embeddings.
\newblock In {\em Proceedings of the 2018 Conference of the North American
  Chapter of the Association for Computational Linguistics: Human Language
  Technologies, Volume 2 (Short Papers)}, pages 175--180, 2018.

\bibitem{shi2010corpus}
Shuming Shi, Huibin Zhang, Xiaojie Yuan, and Ji-Rong Wen.
\newblock Corpus-based semantic class mining: distributional vs. pattern-based
  approaches.
\newblock In {\em Proceedings of the 23rd International Conference on
  Computational Linguistics (Coling 2010)}, pages 993--1001, 2010.

\bibitem{xu2020clue}
Liang Xu, Xuanwei Zhang, Lu~Li, Hai Hu, Chenjie Cao, Weitang Liu, Junyi Li,
  Yudong Li, Kai Sun, Yechen Xu, et~al.
\newblock Clue: A chinese language understanding evaluation benchmark.
\newblock {\em arXiv preprint arXiv:2004.05986}, 2020.

\bibitem{bollacker2008freebase}
Kurt Bollacker, Colin Evans, Praveen Paritosh, Tim Sturge, and Jamie Taylor.
\newblock Freebase: a collaboratively created graph database for structuring
  human knowledge.
\newblock In {\em Proceedings of the 2008 ACM SIGMOD international conference
  on Management of data}, pages 1247--1250, 2008.

\bibitem{ling2012fine}
Xiao Ling and Daniel~S Weld.
\newblock Fine-grained entity recognition.
\newblock In {\em AAAI}, volume~12, pages 94--100, 2012.

\bibitem{weischedel2013ontonotes}
Ralph Weischedel, Martha Palmer, Mitchell Marcus, Eduard Hovy, Sameer Pradhan,
  Lance Ramshaw, Nianwen Xue, Ann Taylor, Jeff Kaufman, Michelle Franchini,
  et~al.
\newblock Ontonotes release 5.0 ldc2013t19.
\newblock {\em Linguistic Data Consortium, Philadelphia, PA}, 23, 2013.

\bibitem{shi2015automatically}
Shuming Shi, Yuehui Wang, Chin-Yew Lin, Xiaojiang Liu, and Yong Rui.
\newblock Automatically solving number word problems by semantic parsing and
  reasoning.
\newblock In {\em Proceedings of the 2015 Conference on Empirical Methods in
  Natural Language Processing}, pages 1132--1142, 2015.

\bibitem{earley1970efficient}
Jay Earley.
\newblock An efficient context-free parsing algorithm.
\newblock {\em Communications of the ACM}, 13(2):94--102, 1970.

\bibitem{xue2005penn}
Naiwen Xue, Fei Xia, Fu-Dong Chiou, and Marta Palmer.
\newblock The penn chinese treebank: Phrase structure annotation of a large
  corpus.
\newblock {\em Natural language engineering}, 11(2):207, 2005.

\bibitem{Santorini1990PartofSpeechTG}
Beatrice Santorini.
\newblock Part-of-speech tagging guidelines for the penn treebank project (3rd
  revision).
\newblock 1990.

\bibitem{ratnaparkhi1996maximum}
Adwait Ratnaparkhi.
\newblock A maximum entropy model for part-of-speech tagging.
\newblock In {\em Conference on empirical methods in natural language
  processing}, 1996.

\bibitem{lafferty2001conditional}
John Lafferty, Andrew McCallum, and Fernando~CN Pereira.
\newblock Conditional random fields: Probabilistic models for segmenting and
  labeling sequence data.
\newblock 2001.

\bibitem{akbik2018coling}
Alan Akbik, Duncan Blythe, and Roland Vollgraf.
\newblock Contextual string embeddings for sequence labeling.
\newblock In {\em {COLING} 2018, 27th International Conference on Computational
  Linguistics}, pages 1638--1649, 2018.

\bibitem{liu2019roberta}
Yinhan Liu, Myle Ott, Naman Goyal, Jingfei Du, Mandar Joshi, Danqi Chen, Omer
  Levy, Mike Lewis, Luke Zettlemoyer, and Veselin Stoyanov.
\newblock Roberta: A robustly optimized bert pretraining approach.
\newblock {\em arXiv preprint arXiv:1907.11692}, 2019.

\bibitem{li2020segmenting}
Yangming Li, Lemao Liu, and Shuming Shi.
\newblock Segmenting natural language sentences via lexical unit analysis.
\newblock {\em arXiv preprint arXiv:2012.05418}, 2020.

\bibitem{Kitaev2018ConstituencyPW}
Nikita Kitaev and D.~Klein.
\newblock Constituency parsing with a self-attentive encoder.
\newblock In {\em ACL}, 2018.

\bibitem{shi2019simple}
Peng Shi and Jimmy Lin.
\newblock Simple bert models for relation extraction and semantic role
  labeling.
\newblock {\em arXiv preprint arXiv:1904.05255}, 2019.

\bibitem{Chen2017EnhancedLF}
Qian Chen, Xiao-Dan Zhu, Z.~Ling, Si~Wei, Hui Jiang, and Diana Inkpen.
\newblock Enhanced lstm for natural language inference.
\newblock In {\em ACL}, 2017.

\bibitem{li2020empirical}
Yangming Li, Lemao Liu, and Shuming Shi.
\newblock Empirical analysis of unlabeled entity problem in named entity
  recognition.
\newblock {\em arXiv preprint arXiv:2012.05426}, 2020.

\bibitem{gillick2014context}
Dan Gillick, Nevena Lazic, Kuzman Ganchev, Jesse Kirchner, and David Huynh.
\newblock Context-dependent fine-grained entity type tagging.
\newblock {\em arXiv preprint arXiv:1412.1820}, 2014.

\bibitem{sang2003introduction}
Erik~F Sang and Fien De~Meulder.
\newblock Introduction to the conll-2003 shared task: Language-independent
  named entity recognition.
\newblock {\em arXiv preprint cs/0306050}, 2003.

\bibitem{derczynski2016broad}
Leon Derczynski, Kalina Bontcheva, and Ian Roberts.
\newblock Broad twitter corpus: A diverse named entity recognition resource.
\newblock In {\em Proceedings of COLING 2016, the 26th International Conference
  on Computational Linguistics: Technical Papers}, pages 1169--1179, 2016.

\bibitem{Bos2017GMB}
Johan Bos, Valerio Basile, Kilian Evang, Noortje Venhuizen, and Johannes
  Bjerva.
\newblock The groningen meaning bank.
\newblock In Nancy Ide and James Pustejovsky, editors, {\em Handbook of
  Linguistic Annotation}, volume~2, pages 463--496. Springer, 2017.

\bibitem{alvarado2015domain}
Julio Cesar~Salinas Alvarado, Karin Verspoor, and Timothy Baldwin.
\newblock Domain adaption of named entity recognition to support credit risk
  assessment.
\newblock In {\em Proceedings of the Australasian Language Technology
  Association Workshop 2015}, pages 84--90, 2015.

\bibitem{balasuriya2009named}
Dominic Balasuriya, Nicky Ringland, Joel Nothman, Tara Murphy, and James~R
  Curran.
\newblock Named entity recognition in wikipedia.
\newblock In {\em Proceedings of the 2009 Workshop on The People’s Web Meets
  NLP: Collaboratively Constructed Semantic Resources (People’s Web)}, pages
  10--18, 2009.

\bibitem{nothman2013learning}
Joel Nothman, Nicky Ringland, Will Radford, Tara Murphy, and James~R Curran.
\newblock Learning multilingual named entity recognition from wikipedia.
\newblock {\em Artificial Intelligence}, 194:151--175, 2013.

\bibitem{derczynski2017results}
Leon Derczynski, Eric Nichols, Marieke van Erp, and Nut Limsopatham.
\newblock Results of the wnut2017 shared task on novel and emerging entity
  recognition.
\newblock In {\em Proceedings of the 3rd Workshop on Noisy User-generated
  Text}, pages 140--147, 2017.

\bibitem{marcus1993building}
Mitchell Marcus, Beatrice Santorini, and Mary~Ann Marcinkiewicz.
\newblock Building a large annotated corpus of english: The penn treebank.
\newblock 1993.

\bibitem{pradhan2013towards}
Sameer Pradhan, Alessandro Moschitti, Nianwen Xue, Hwee~Tou Ng, Anders
  Bj{\"o}rkelund, Olga Uryupina, Yuchen Zhang, and Zhi Zhong.
\newblock Towards robust linguistic analysis using ontonotes.
\newblock In {\em Proceedings of the Seventeenth Conference on Computational
  Natural Language Learning}, pages 143--152, 2013.

\bibitem{liu-etal-2018-lcqmc}
Xin Liu, Qingcai Chen, Chong Deng, Huajun Zeng, Jing Chen, Dongfang Li, and
  Buzhou Tang.
\newblock {LCQMC}:a large-scale {C}hinese question matching corpus.
\newblock In {\em Proceedings of the 27th International Conference on
  Computational Linguistics}, August 2018.

\bibitem{chen-etal-2018-bq}
Jing Chen, Qingcai Chen, Xin Liu, Haijun Yang, Daohe Lu, and Buzhou Tang.
\newblock The {BQ} corpus: A large-scale domain-specific {C}hinese corpus for
  sentence semantic equivalence identification.
\newblock In {\em Proceedings of the 2018 Conference on Empirical Methods in
  Natural Language Processing}, October-November 2018.

\bibitem{zhang+:2018:paws}
Yuan Zhang, Jason Baldridge, and Luheng He.
\newblock {PAWS:} paraphrase adversaries from word scrambling.
\newblock {\em CoRR}, abs/1904.01130, 2019.

\bibitem{hinton2015distilling}
Geoffrey Hinton, Oriol Vinyals, and Jeff Dean.
\newblock Distilling the knowledge in a neural network.
\newblock {\em arXiv preprint arXiv:1503.02531}, 2015.

\end{thebibliography}
\bibliographystyle{arxiv.sty}

\end{CJK}
\end{document}